\documentclass[letterpaper, 10 pt, conference]{ieeeconf}  

\usepackage{times}

\usepackage{multicol}

\usepackage[utf8]{inputenc} 
\usepackage[T1]{fontenc}    
\usepackage{booktabs}       
\usepackage{amsfonts}       
\usepackage{nicefrac}       
\usepackage{microtype}  
\usepackage{makecell}
\usepackage{capt-of}

\usepackage{wrapfig}

\usepackage{pgfplots}
\usepackage{pgfplotstable}
\pgfplotsset{compat=1.18}
\usepgfplotslibrary{colorbrewer}
\usetikzlibrary{patterns}

\usepackage[backref=page]{hyperref}

\makeatletter
\let\NAT@parse\undefined
\makeatother
\usepackage[numbers]{natbib}

\usepackage{amsmath,amssymb}
\usepackage{graphicx}
\usepackage{bbm}
\usepackage{subcaption}
\usepackage[font=small]{caption}

\let\labelindent\relax
\usepackage{enumitem}
\usepackage{algorithm}
\usepackage[noend]{algpseudocode}
\usepackage{colortbl}
\usepackage{xcolor}
\usepackage{multirow}

\usepackage{titletoc}
\usepackage[T1]{fontenc}    
\usepackage{url}            

\usepackage{cleveref}

\usepackage{bbm}

\usepackage{siunitx}
\usepackage{tablefootnote}
\usepackage{makecell}
\usepackage{dsfont}
\usepackage{diagbox}
\usepackage[margin=1in]{geometry}

\definecolor{resocolor}{RGB}{231,239,237}
\definecolor{robust6}{RGB}{222,236,242}
\definecolor{imagenet6}{RGB}{220,221,233}

\IEEEoverridecommandlockouts                              

\title{\LARGE \bf
End-to-end RL Improves Dexterous Grasping Policies
}

\author{Ritvik Singh$^{1,2}$, Karl Van Wyk$^{1}$, Jitendra Malik$^{2}$, Pieter Abbeel$^{2}$, Nathan Ratliff$^{1}$, Ankur Handa$^{1}$
\thanks{$^{1}$NVIDIA}%
\thanks{$^{2}$University of California, Berkeley}%
}

\begin{document}

\IEEEaftertitletext{%
\noindent\begin{minipage}{\textwidth}\centering
\begin{tabular}{@{}c@{\hspace{2pt}}c@{\hspace{2pt}}c@{\hspace{2pt}}c@{}}
\includegraphics[width=.25\linewidth]{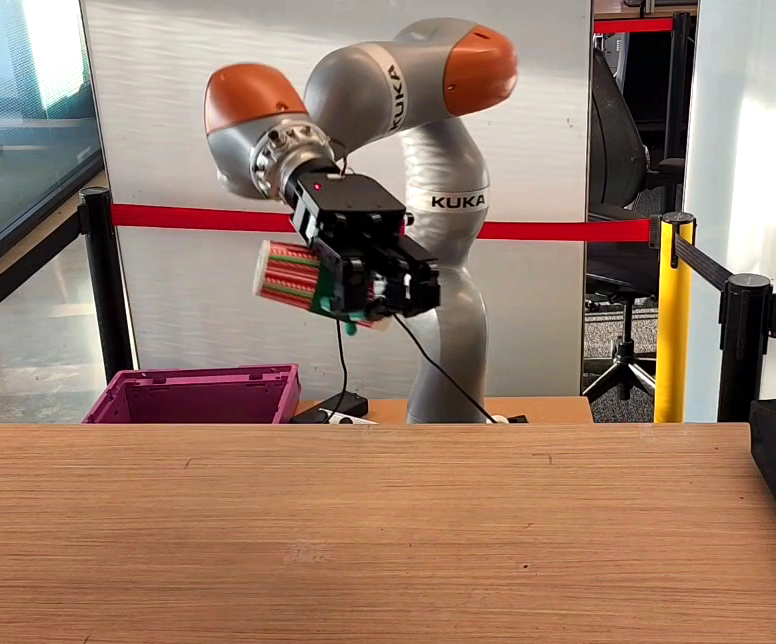} &
\includegraphics[width=.25\linewidth]{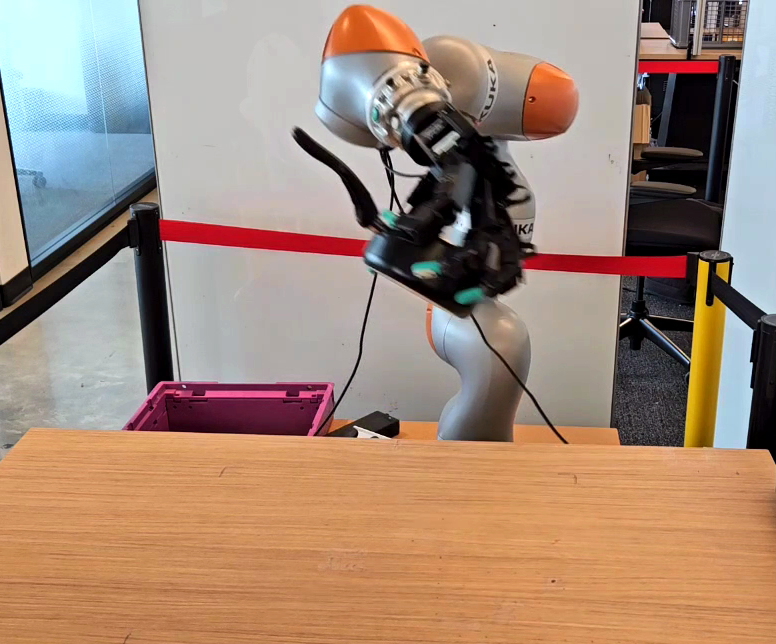} &
\includegraphics[width=.25\linewidth]{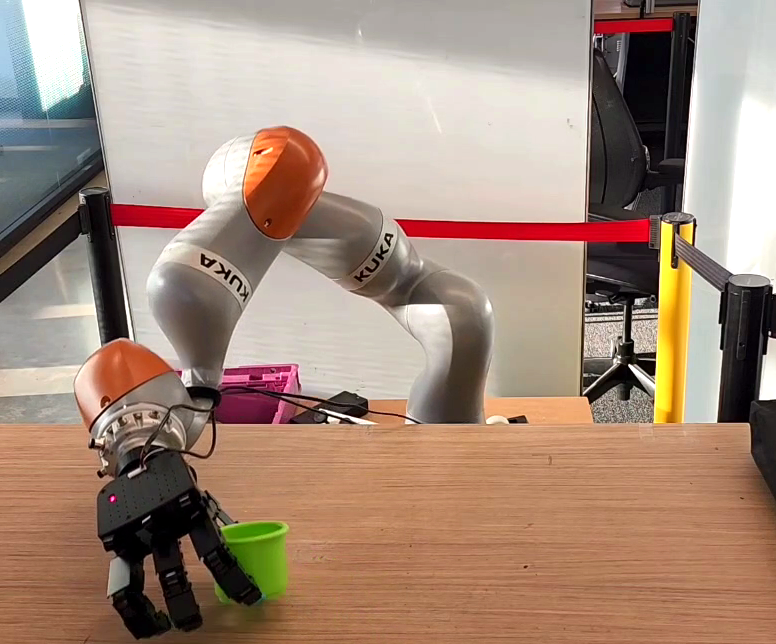} &
\includegraphics[width=.25\linewidth]{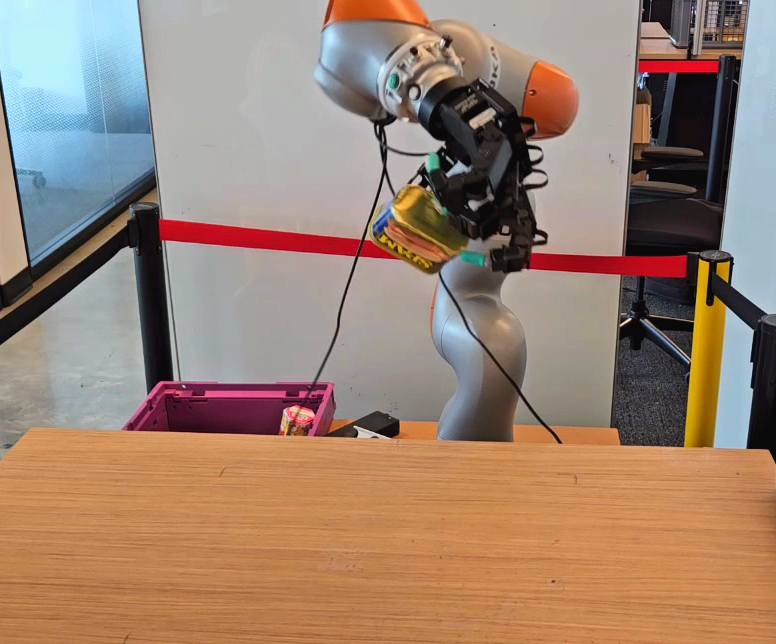} \\[-2pt]
\includegraphics[width=.25\linewidth]{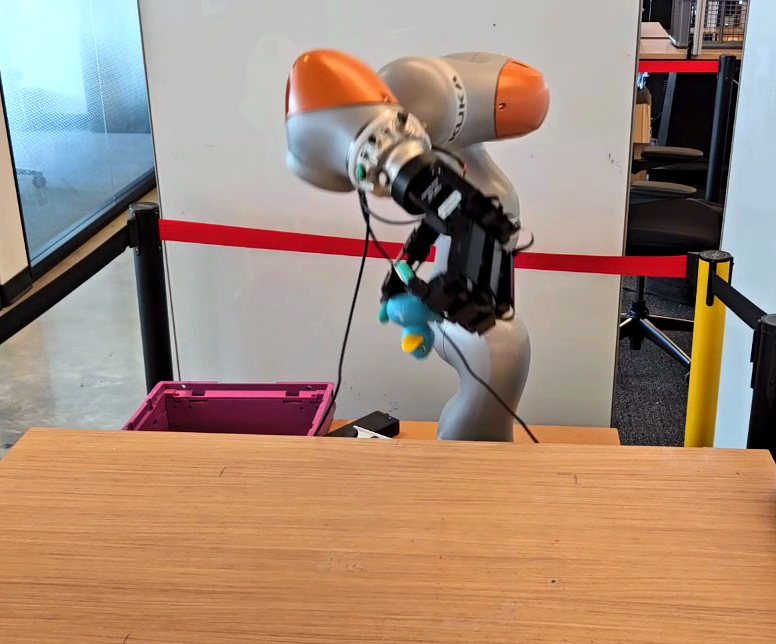} &
\includegraphics[width=.25\linewidth]{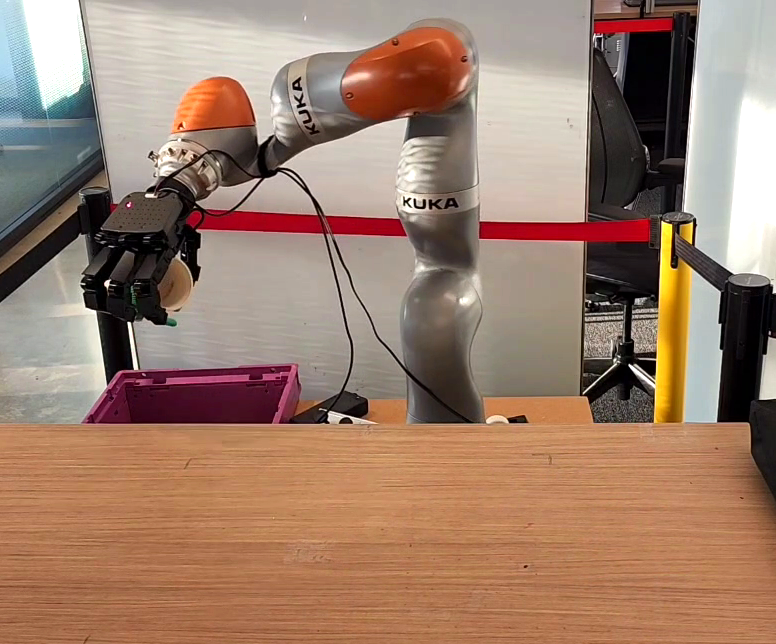} &
\includegraphics[width=.25\linewidth]{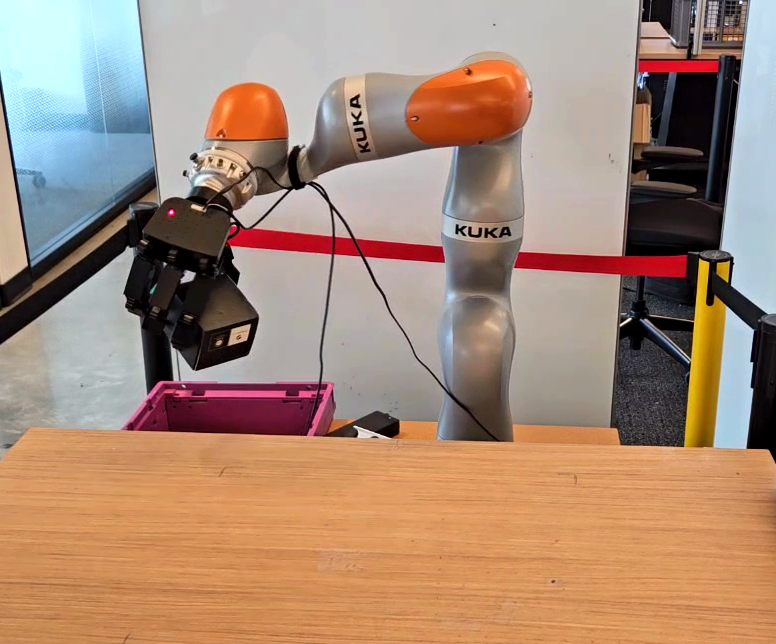} &
\includegraphics[width=.25\linewidth]{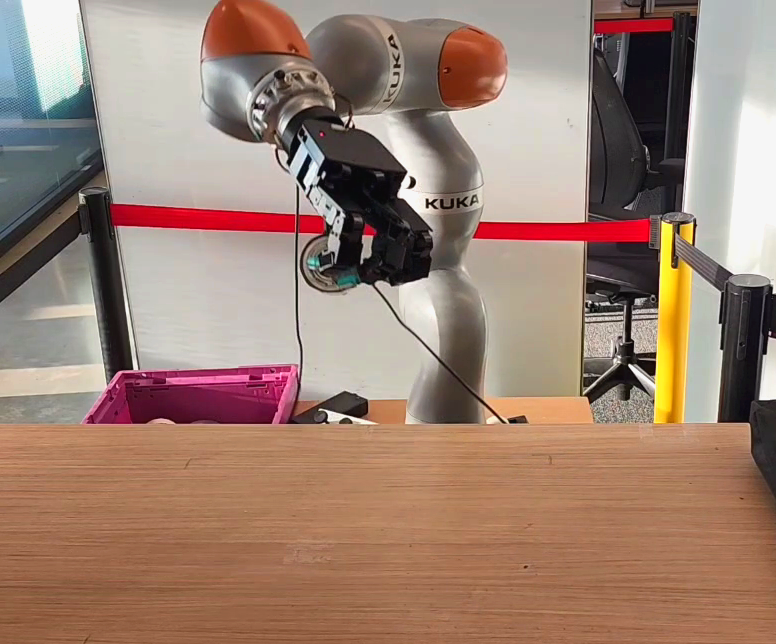}
\end{tabular}
\captionof{figure}{We deploy our policy in the real world with varying lighting conditions. All of the objects in the above figure were unseen during training.}
\label{fig:title_fig}
\end{minipage}%
}

\maketitle

\begin{abstract}
This work explores techniques to scale up image-based end-to-end learning for dexterous grasping with an arm + hand system. Unlike state-based RL, vision-based RL is much more memory inefficient, resulting in relatively low batch sizes, which is not amenable for algorithms like PPO. Nevertheless, it is still an attractive method as unlike the more commonly used techniques which distill state-based policies into vision networks, end-to-end RL can allow for emergent active vision behaviors. We identify a key bottleneck in training these policies is the way most existing simulators scale to multiple GPUs using traditional data parallelism techniques. We propose a new method where we disaggregate the simulator and RL (both training and experience buffers) onto separate GPUs. On a node with four GPUs, we have the simulator running on three of them, and PPO running on the fourth. We are able to show that with the same number of GPUs, we can double the number of existing environments compared to the previous baseline of standard data parallelism. This allows us to train vision-based environments, end-to-end with depth, which were previously performing far worse with the baseline. We train and distill both depth and state-based policies into stereo RGB networks and show that depth distillation leads to better results, both in simulation and reality. This improvement is likely due to the observability gap between state and vision policies which does not exist when distilling depth policies into stereo RGB. We further show that the increased batch size brought about by disaggregated simulation also improves real world performance. When deploying in the real world, we improve upon the previous state-of-the-art vision-based results using our end-to-end policies. More information can be found at \url{https://e2e4robotics.com/}.
\end{abstract}

\section{INTRODUCTION}

In robotics, crafting behaviours which exhibit a mix of agility, reactivity, and dexterity when interacting with the environment remains a longstanding goal. Recently, reinforcement  and imitation learning have emerged as powerful paradigms for the training of robot policies. For the vast majority of everyday tasks, endowing policies with vision is a required component to sense the environment and achieve the desired behaviour. In recent years, visuomotor policy learning - learning policies which take in a mix of visual and proprioceptive inputs - has emerged as a powerful paradigm for the creation of robotic policies for a wide variety of tasks \citep{chi2024universalmanipulationinterfaceinthewild, zhao2023learningfinegrainedbimanualmanipulation}. As part of this, so-called end-to-end methods of direct processing of image to action have emerged as a useful technique for representing policies, sidestepping the need for explicit representations of the environment state \citep{levine2016endtoendtrainingdeepvisuomotor}. Such methods usually take in raw RGB or depth camera observations, and produce an action output that is the result of processing using a neural network.

One popular approach to producing visual policies is via distillation in simulation \citep{Lee_2020, Miki_2022, kumar2021rmarapidmotoradaptation, agarwal2022leggedlocomotionchallengingterrains,lum2024dextrahgpixelstoactiondexterousarmhand, singh2025dextrahrgbvisuomotorpoliciesgrasp}. In such approaches, an expert policy, commonly known as the "teacher", is learned, usually via Reinforcement Learning (RL), and then a downstream "student" policy is learned by rolling out the policy and supervising its actions based on the expert policy's output. The advantage of this method is that the student and the teacher do not need to have the same inputs, allowing the practitioner great flexibility in designing the input space and network architecture for both the teacher RL training (where it is possible to include privileged information only available in the simulator, e.g. ground truth states) and in the student during distillation. This way, the sample inefficiency of RL is limited to just the privileged teacher which is much easier than visual RL where the state space is far more complex. This leads to a "factorization" of the visuomotor learning process into behavior learning in the first stage and representation learning in the second. While convenient, this approach has several limitations. Namely, it results in vision-based policies that learn state-based behaviors. This is problematic because this can create a partial observability problem whereby the student has difficulty replicating the teacher actions.


A solution to this would be to simply train the vision policy with RL. While this has been employed for primitive grasping tasks~\cite{levine2016endtoendtrainingdeepvisuomotor, james2019simtorealsimtosimdataefficientrobotic}, it has yet to be done for complex, dexterous manipulation tasks as the sample complexity in pixel space is a lot higher. This requires scaling up the number of simulated environments. However, simulating batched rendering environments in parallel is very memory intensive. This means that without access to a large number of GPUs, it becomes difficult to scale up the number of environments in order to train such vision policies from scratch with RL. In order to more efficiently utilize GPU memory, we take inspiration from the disaggregated prefill and decode setups that are used for modern day LLM inferencing~\cite{zhong2024distservedisaggregatingprefilldecoding}. We introduce a disaggregated simulation and RL framework, which separates RL experience buffers/training and simulation environments onto separate GPUs. This allows us to simulate twice as many environments compared to the previous data parallelism baseline on the exact same hardware setup. With this, we can train end-to-end depth policies for a Kuka-Allegro robot setup to perform the task of dexterous grasping. We distill these depth policies into RGB policies which avoids the teacher-student distillation gap previously seen with state-based teachers. Ultimately, we find that in both simulation and the real world, policies distilled from vision-based teachers are more performant than policies distilled from state-based ones. Figure~\ref{fig:title_fig} shows snippets of the robot grasping various objects unseen during training.

\section{Related Work}
\textbf{Scaling frameworks for vision-based RL.} There are many existing simulators for robot learning that support rendering such as Isaac Gym~\cite{makoviychuk2021isaacgymhighperformance}, Isaac Lab~\cite{mittal2023orbit}, ManiSkill~\cite{tao2025maniskill3gpuparallelizedrobotics}, and MuJoCo~\cite{zakka2025mujocoplayground}. However, the current paradigm with which they scale up RL is by using data parallelism, which can be inefficient when it comes to memory use as shown in Section~\ref{sec:disagg_sim}. Impala~\cite{espeholt2018impalascalabledistributeddeeprl} is a scalable distributed RL framework which hosts the learner on a separate machine from the environment. The limitation with this approach is that they also store the trajectory on the same device that contains the environment which becomes expensive for vectorized simulators. When applied to modern robotics simulators, this would result in large experience buffers that cannot fit on the same GPU without lowering the number of environments. SEED RL~\cite{espeholt2020seedrlscalableefficient} is a follow up work which stores the trajectory on the same device as the learner. However, their implementation is not tailored towards modern robotics simulators which are vectorized for GPUs as they built their distributed RL infrastructure around large CPU clusters. Concretely, this means that their framework is used in scenarios with very large number of CPUs, each one simulating a few environments, whereas our method for disaggregated simulation is meant to run on GPU-accelerated simulators which require few replicas as each GPU can simulate many more environments. Lastly, they also do not demonstrate end-to-end learning on robotics tasks that transfer to the real world.

\textbf{Vision Based Grasping with Hands.} There have been several prior works for vision-based grasping with multi-fingered hands. They can either learn from depth/pointclouds~\cite{lum2024dextrahgpixelstoactiondexterousarmhand,zhang2025robustdexgrasprobustdexterousgrasping,lin2025simtorealreinforcementlearningvisionbased} or through RGB~\cite{singh2025dextrahrgbvisuomotorpoliciesgrasp}. While the various implementation details differ among the papers, the one common aspect is that they all leverage a teacher-student distillation pipeline. The teacher policies all operate on privileged information which is then distilled into vision policies that operate in the real world. This means that vision policies are trained to mimic the behavior of privileged policies, which can ultimately lead to suboptimal policy performance. This is because the inability to reproduce teacher actions can result in trajectories that diverge from those of the teacher, which results in distributional shifts that can lead to unrecoverable failures. Although there has been previous work aiming to reduce this partial observability gap~\cite{2014-mfcgps, kim2025distillingrealizablestudentsunrealizable}, there has not been much focus investigating this for complex environments such as vision-based dexterous grasping.

\textbf{End-to-End Visuomotor Policies for Parallel Jaw Grippers.} Several works in the past have demonstrated end-to-end grasping from RL using simple parallel jaw grippers~\cite{james2017transferringendtoendvisuomotorcontrol, matas2018simtorealreinforcementlearningdeformable, levine2016endtoendtrainingdeepvisuomotor, james2019simtorealsimtosimdataefficientrobotic, kalashnikov2018qtoptscalabledeepreinforcement}. The reduced action space brought by parallel jaw grippers makes the RL process much easier than using hands. However, this morphology can be limiting because they cannot achieve the same types of stable grasps as multifingered hands~\cite{lin2024learningvisuotactileskillsmultifingered}.

\section{Method}
\subsection{End-to-End RL}
Vision-based policies are incredibly important for real-world manipulation. However, training them directly from RL has historically been challenging due to the high sample complexity of image space. This has led to two-stage methods gaining prominence whereby a state-based teacher policy is trained with RL and then distilled into a vision-based student policy. While this has led to training successful policies, they fundamentally do not learn vision-aware behaviors. For example, imagine a robot arm trying to grasp an object. Suppose the arm is currently occluding said object. The teacher policy, which has access to the groundtruth object position, will simply just pick it up. However, the student policy may struggle to recreate this behavior as it has not learned to move the arm out of the way in order to see the object. In such cases, the student is trying to mimic state-based behaviors while only having access to vision information which causes it to act sub-optimally with respect to its inputs. Therefore, end-to-end training, where the RL policy learns directly from images, will lead to policy behaviors that lend themselves better to their sensory modalities.
However, RGB end-to-end RL is much slower than depth-based RL as the rendering process for accurate light transport simulation is far more time consuming. Thus, a suitable middle ground that meets the requirements above while also being able to run on a reasonable hardware budget is to train a depth based policy with RL, and distill this into a stereo RGB-based policy in order to deploy in the real world. This way, there is no theoretical information gap between the student and the teacher.

\subsection{Disaggregated Simulation and RL} \label{sec:disagg_sim}
\begin{figure}[htbp]  
  \centering
  \begin{subfigure}[b]{0.49\textwidth}
    \centering
    \includegraphics[width=\linewidth]{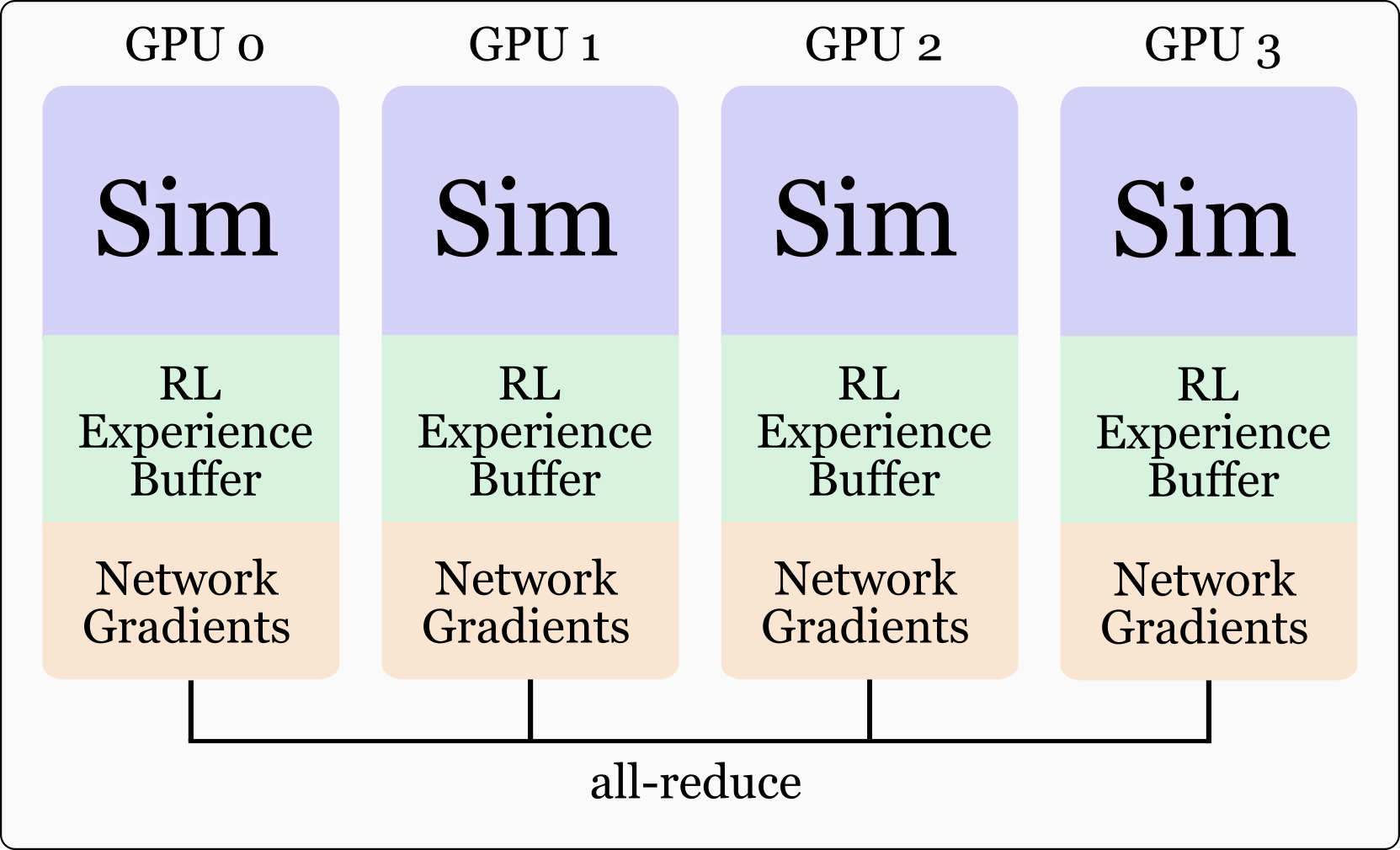}   
    \caption{Traditional data-parallel setup.}
    \label{fig:sim_dp}
  \end{subfigure}\hfill
  \begin{subfigure}[b]{0.49\textwidth}
    \centering
    \includegraphics[width=\linewidth]{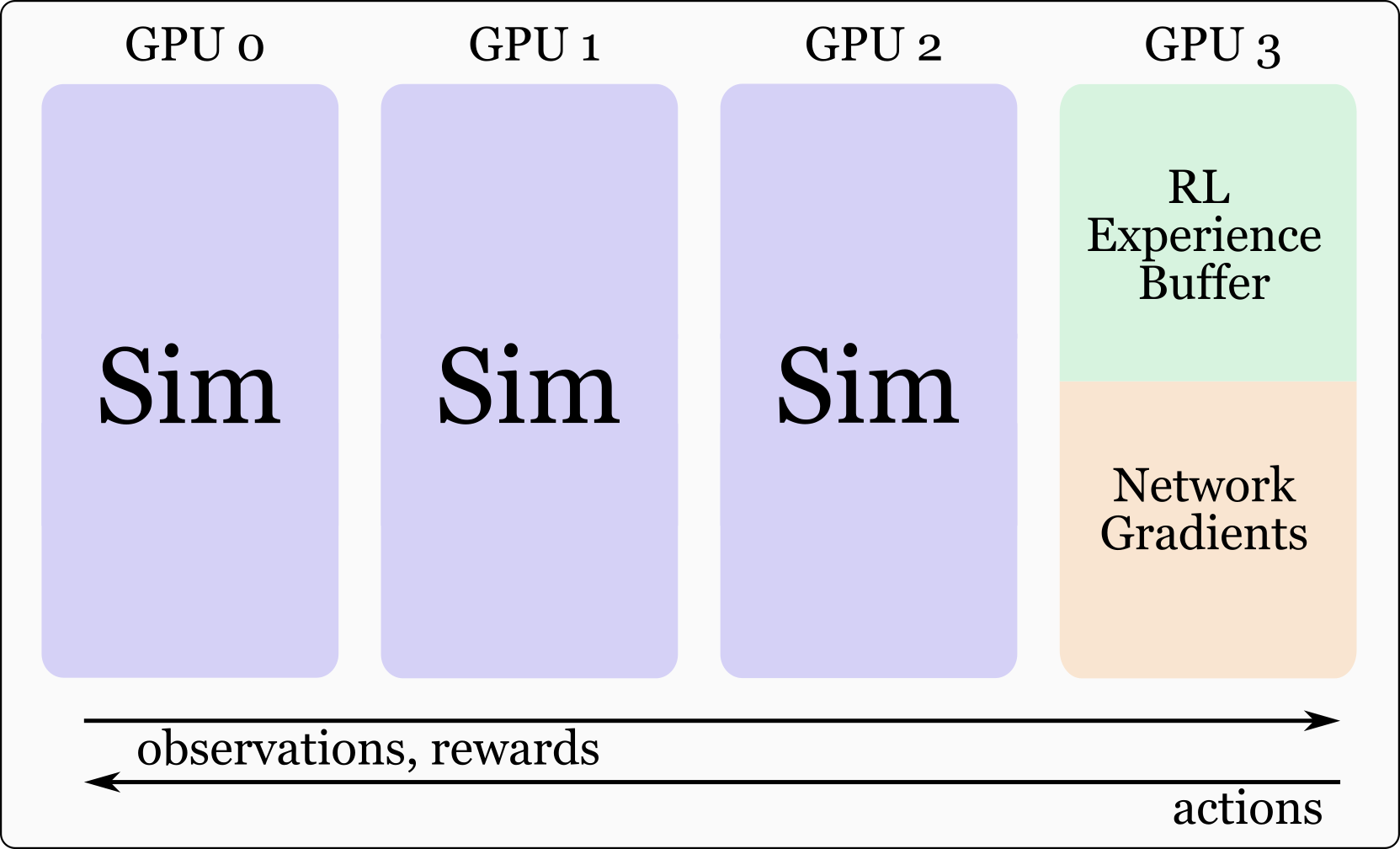}
    \caption{Disaggregated RL setup.}
    \label{fig:sim_disagg}
  \end{subfigure}
  \caption{Disaggregated simulation/learning pipeline.  %
           (a) The standard data parallel setup when scaling up where every GPU runs both simulation and RL;  %
           (b) our disaggregated simulation setup where three GPUs purely run training and the last one stores the experience buffers and runs RL.}
  \label{fig:simulation_setups}
\end{figure}

When training end-to-end RL, it is important to scale up the number of environments in order to get a reliable learning signal. Scaling up reinforcement learning is typically done through naive data parallelism. This means that every GPU runs the simulator, stores the RL experience buffers, and computes the gradients for the actor and critic. After each instance calculates the gradients, they are averaged across all GPUs in order to obtain a less-noisy gradient estimate (see Fig \ref{fig:sim_dp}). When performing end-to-end vision RL, the experience buffers can balloon in size. For example, assuming depth maps are represented using standard fp32 precision, storing the experience across just 512 environments at 320x240 resolution for a relatively short horizon length of 16 steps can take up to 2.5GB alone. The simulator will also start to consume more memory as more environments are added. Crucially, however; the simulator will take up a non-trivial amount of memory that is not a function of the number of environments which is typically used for the asset cache. Therefore, naively running the simulator on every GPU is a suboptimal use of memory as each copy will bring with it the same large asset cache. 

A more optimal use of memory would involve limiting the amount of GPUs that run the simulation, and ensuring that the ones that are running simulation do not have to also store RL experience buffers or network gradients. We present our proposed disaggregated simulation and RL setup in Fig \ref{fig:sim_disagg}. In a node of 4 GPUs, we run the simulator on 3 of them, in order to fully maximize the number of environments we can simulate, and on the 4th GPU, we run RL training and store the experience buffers. Psuedocode for how the different GPUs communicate with each other is shown in Algorithms~\ref{alg:sim-simple} and~\ref{alg:learner-simple}.

It is important to maximize the number of environments because unlike traditional state-based environments which can be scaled up to the point of PPO saturation long before the GPU is out of memory, vision environments can be far more memory intensive. For example, the state-based DextrAH task~\cite{lum2024dextrahgpixelstoactiondexterousarmhand, singh2025dextrahrgbvisuomotorpoliciesgrasp}, which involves object grasping with a Kuka and Allegro hand-arm system, can be reliably solved with 4096 environments which only takes up 14GB of memory. In contrast, the same task, when using 320x240 depth cameras, will take up 44GB just to simulate 256 environments. Thus, maximizing the number of environments becomes crucial when trying to solve vision tasks with RL. Our disaggregated method, as shown in Table~\ref{tab:num_envs}, is able to double the number of environments that are able to be simulated on the exact same hardware.

\begin{table}[h]
  \setlength{\tabcolsep}{4pt} 
  \centering
  \caption{Maximum concurrent environments on one 4-GPU NVIDIA L40S node at various resolutions. Disaggregated Simulation is able to more than double the number of simulated environments on the same hardware compared to traditional data parallelism.}
  \label{tab:num_envs}
  \begin{tabular}{@{}lcc@{}}
    \toprule
    \textbf{Input resolution} & \textbf{Data Parallel} & \textbf{Disaggregated Simulation} \\
    \midrule
    $160\times120$ &
      \makecell{1024 / GPU\\(4096 total)} &
      \makecell{2800 / GPU\\(8400 total)} \\[2pt]
    $320\times240$ &
      \makecell{256 / GPU\\(1024 total)} &
      \makecell{700 / GPU\\(2100 total)} \\
    \bottomrule
  \end{tabular}
  \footnotetext[1]{4×“sim + RL”.}
  \footnotetext[2]{3×simulators + 1×learner.}
\end{table}

\begin{algorithm}[t]
\caption{Simulation Replica (GPU $s \in \{0,1,2\}$)}
\label{alg:sim-simple}
\begin{algorithmic}[1]
\State \textbf{Given:} learner id $\ell \gets 3$, environment $\mathsf{Env}$
\State obs $\gets \mathsf{Env.Reset}()$
\State \Call{SendTo}{$\ell$, obs} \Comment{send initial obs to learner}
\While{true}
  \State actions $\gets$ \Call{Recv}{$\ell$}
  \State $(\text{obs}, \text{rew}, \text{dones}) \gets \mathsf{Env.Step}(\text{actions})$
  \State \Call{SendTo}{$\ell$, (rew, dones, obs)}
\EndWhile
\end{algorithmic}
\end{algorithm}

\begin{algorithm}[t]
\caption{Learner / Trainer (GPU $3$)}
\label{alg:learner-simple}
\begin{algorithmic}[1]
\State \textbf{Given:} sim ids $\mathcal{S} \gets \{0,1,2\}$, horizon $H$, policy $\pi$
\ForAll{$s \in \mathcal{S}$} \State obs[$s$] $\gets$ \Call{Recv}{$s$} \EndFor \Comment{initial obs}
\While{true}
  \State $\mathcal{D} \gets \emptyset$ \Comment{trajectory buffer}
  \For{$t = 1 \ \textbf{to}\ H$}
     \State actions $\gets \pi\big(\text{stack}(\{\text{obs}[s]\}_{s\in\mathcal{S}})\big)$
     \ForAll{$s \in \mathcal{S}$}
        \State \Call{SendTo}{$s$, \,actions[$s$]} 
        \State rew[$s$], dones[$s$], nextObs[$s$] $\gets$ \Call{Recv}{$s$}
     \EndFor
     \State $\mathcal{D} \gets D \bigcup \{($obs, actions, rew, done$)\}$
     \State obs $\gets$ nextObs
  \EndFor
  \State \Call{TrainPPO}{$\pi, \mathcal{D}$}
\EndWhile
\end{algorithmic}
\end{algorithm}

\subsection{Training Environment}\label{sec:training_env}
    We use the same environment as in DextrAH-RGB~\cite{singh2025dextrahrgbvisuomotorpoliciesgrasp} and provide a brief summary of it here. The environment consists of a 7 DoF Kuka iiwa arm and a 16 DoF Allegro V4 hand. The policy is trained in simulation to grasp and lift 140 different objects from the Visual Dexterity dataset~\cite{Chen_2023}. To help facilitate sim-to-real transfer, we employ domain randomization on a set of physics parameters such as joint friction, stiffness, damping, mass, etc. This is implemented through a method known as Automatic Domain Randomization (ADR) ~\cite{openai2019solvingrubikscuberobot, singh2025dextrahrgbvisuomotorpoliciesgrasp} which sets an initial randomization range for the various physics parameters and gradually increases the range towards the terminal range as the policy becomes more proficient at grasping and lifting the object. This is done to induce a curriculum onto the environment that balances exploration early on in training while ensuring robustness of the policy.

\begin{figure}[htbp]
    \centering
    \includegraphics[width=0.5\textwidth]{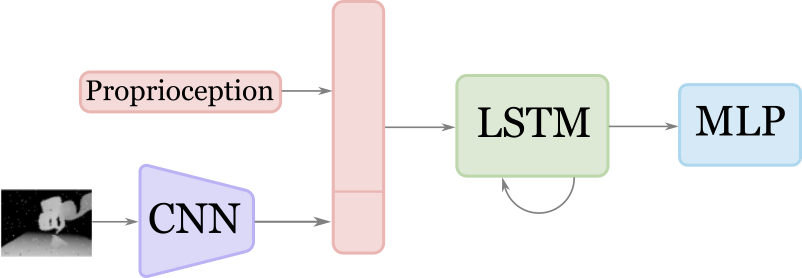}%
    \caption{The image is first passed into a 4-layer CNN which outputs a 32-dimensional embedding. This is concatenated with the rest of the robot proprioceptive data and then fed through to the LSTM and then MLP.}
    \label{fig:depth_arch}
    \end{figure}

    We train our depth-based policies end-to-end with PPO~\cite{schulman2017proximalpolicyoptimizationalgorithms}. Our network architecture comprises a 4-layer CNN with $[16, 32, 64, 128]$ filters, layer normalization, and ReLU activation function. The output of this is passed into a fully connected layer which outputs a 32 dimensional embedding for the depth. This is combined with the proprioception of the robot and fed into two LSTM layers with $1024$ units, the output of which is fed into a 3 layer fully-connected network with $[512, 512, 256]$ hidden units. This is in contrast to IMPALA~\cite{espeholt2018impalascalabledistributeddeeprl} where the fully connected network is placed before the LSTM. The architecture for the policy is shown in Figure~\ref{fig:depth_arch}.

\begin{figure}[htbp]
    \centering
    \includegraphics[width=0.485\textwidth]{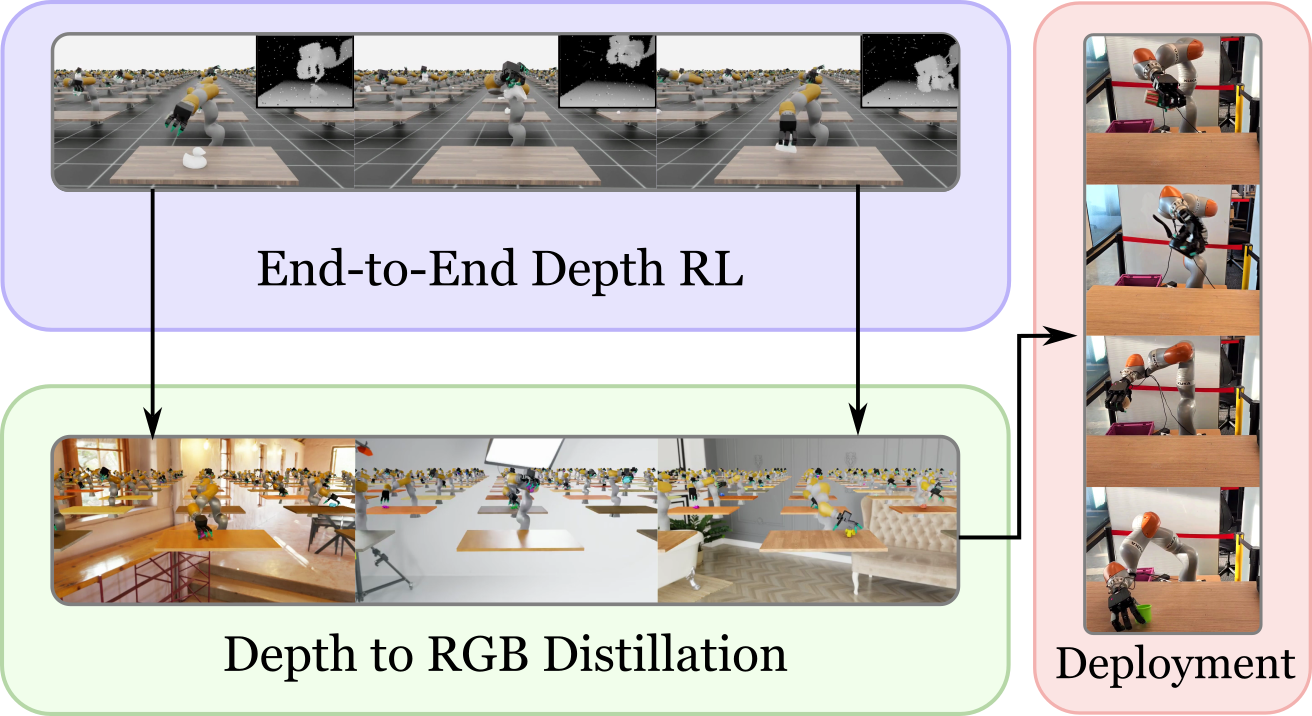}%
    \caption{We first train our depth-based teacher policies end-to-end using reinforcement learning. We then distill these depth policies into stereo RGB. Lastly, these policies are then deployed into the real world.}
    \label{fig:pipeline}
    \end{figure}

    We choose to not train direct end-to-end RGB policies as realistic RGB rendering can be significantly more time consuming than simple tiled depth rendering. However, we still want to retain the benefits of RGB over depth as shown in~\cite{singh2025dextrahrgbvisuomotorpoliciesgrasp}. Thus, we chose to distill the depth policy into a stereo RGB policy with the same architecture as~\cite{singh2025dextrahrgbvisuomotorpoliciesgrasp}. The overall pipeline is shown in Figure~\ref{fig:pipeline}. Because depth is recoverable from a stereo RGB pair, the behaviors learned by a depth policy can theoretically be completely replicated by a distilled stereo RGB policy. This is in contrast with trying to imitate the behavior of a state-based policy which has a much larger observation gap. As shown in Section~\ref{sec:results} our method leads to improved performance in both simulation and reality.

\newcommand{\tablestyle}[2]{\setlength{\tabcolsep}{#1}\renewcommand{\arraystretch}{#2}}

\newcommand{\frozenicon}{\includegraphics[height=1em]{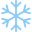}}  

\vspace{-2mm}
\section{Results}\label{sec:results}
We perform various experiments to demonstrate the benefit of end-to-end RL to produce vision-based teachers over the traditional factorization of state-based teachers and vision-based students. We empirically verify both in simulation and in reality the benefit of end-to-end RL on success rates. We further show that our design choice of disaggregated simulation helps improve batch size and thus, downstream performance. All experiments are done on a 4-GPU NVIDIA L40S node.

\subsection{Disaggregated Simulation vs Data Parallelism}
We verify the efficacy of our disaggregated simulation setup by comparing it with data parallelism. For each type of parallelism method, we run experiments at two sets of resolutions: $160 \times 120$ and $320 \times 240$. For each experiment, we run five different seeds and take the average. For each resolution, we use the corresponding number of environments as reported in Table~\ref{tab:num_envs}. The results of this experiment are shown in Table~\ref{tab:dp_vs_disagg}. The first metric we report is the average portion of the terminal domain randomization (DR) ranges that was achieved by ADR. Since ADR increases the DR range towards the terminal range when the success rate is at least $0.4$, a higher value indicates that the policy was more performant, causing ADR to increase the ranges more. The second metric we track is the percentage of runs that managed to reach the terminal ADR range. This metric measures how reliable the training setup is in producing performant policies. The last metric that we report is the average success rate for all seeds that reached the terminal ADR range. The success rate is defined as the percentage of objects that have been grasped. After being held in the air for 10 seconds, the environments will then reset. For the $320\times240$ data parallel case, none of the seeds reached the terminal DR range, so we report zero.

We see that across both resolutions, our disaggregated simulation setup is able to offer superior performance across all metrics when compared to the standard data-parallelism solution that the other simulators utilize. For the $160\times120$ resolution, not only are runs more likely to reach their terminal DR ranges, but once they do, they also display better grasping performance. For $320\times240$, it was impossible to train any policy to grasp the objects when using data parallelism. This is likely because the number of simulated environments goes down by a factor of four which significantly constrains the PPO batch size. Although our method is unable to reliably train policies to reach their terminal ranges at higher resolutions, it is still able to significantly improve over the baseline. These results show that existing end-to-end tasks are heavily constrained by the number of environments that can be simulated. Thus, our method, which doubles the number of environments that can be simulated on the same hardware, directly contributes to the improved performance of end-to-end vision-based RL policy training.

\begin{table}[t]
  \setlength{\tabcolsep}{4pt} 
  \centering
  \caption{Data Parallel (DP) vs Disaggregated Simulation (Disagg) at $160\times120$ and $320\times240$ runs averaged across five seeds. The first metric is the portion of the terminal DR ranges that were achieved by ADR. The second metric is percentage of seeds that reached the terminal DR ranges. The last metric is the success rate, which measures the portion of environments that have the object grasped in the air (note: if the object has been grasped for 10 seconds, the environment is reset and the object is placed back onto the table).}
  \label{tab:dp_vs_disagg}
  \small
  \newcommand{\best}[1]{\textbf{#1}}
  \begin{tabular}{@{}llrrr@{}}
    \toprule
    \textbf{Res.} & \textbf{Method} & \textbf{ADR Inc.} ↑ & \textbf{\% Full ADR} ↑ & \textbf{SR} ↑ \\
    \midrule
    \multirow{2}{*}{$160\times120$}
      & DP      & 0.38   & 20\%   & 0.37 \\
      & Disagg  & \best{1.0}   & \best{100\%} & \best{0.42} \\
    \midrule
    \addlinespace[2pt]
    \multirow{2}{*}{$320\times240$}
      & DP      & 0.0    & 0\%    & 0.00 \\
      & Disagg  & \best{0.90} & \best{20\%}  & \best{0.35} \\
    \bottomrule
  \end{tabular}
\end{table}

\subsection{Distilling State Teachers vs Depth Teachers}
\begin{figure}[!t]
\centering
\includegraphics[width=0.5\textwidth]{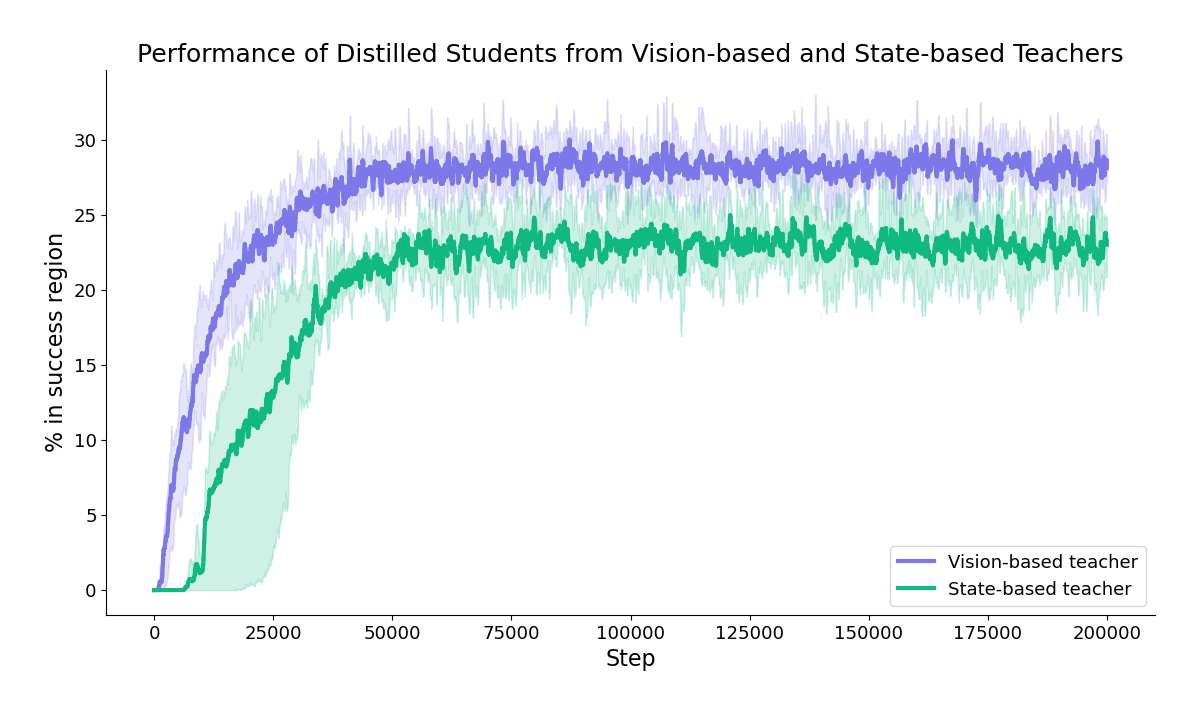}%
\caption{The average performance of RGB students distilled from vision-based teacher policies (blue) and state-based teacher policies (red). It is clear from the plot that students perform better when the teacher policy is also vision-based.}
\label{fig:distillation_perf}
\end{figure}

In order to test the hypothesis that vision teachers offer superior performance to state teachers, we distill both into stereo RGB students to compare the policy performance. For each modality, we train 3 different seeds. The depth teachers were trained as described in Section~\ref{sec:training_env} with disaggregated simulation at $160 \times 120$ resolution to maximize the batch size for RL. The state teachers were trained similar to previous work~\cite{lum2024dextrahgpixelstoactiondexterousarmhand, singh2025dextrahrgbvisuomotorpoliciesgrasp}. In those papers, the state teachers were given access to the object pose and a one-hot vector that corresponds to the object category as the observation. The success metric is the percentage of objects that have been grasped into the air. It is important to note that this is an instantaneous metric and is not the success rate of picking up all objects. Rather, this instantaneously measures how many environments have lifted the object successfully in the air. Once the object has been lifted in the air for two seconds, the environment resets. A higher instantaneous success rate means that the policy is faster and more adept at grasping the object. The results of this experiment are shown in Figure~\ref{fig:distillation_perf}. It is clear from the plot that vision-based teachers lead to better performance than state-based ones. This is because there is less of an information asymmetry between the student and the teacher when it also has to operate on vision input. This leads to the students learning behaviors that are more congruent with their modality. For example, with a depth teacher, the student can learn to better deal with occlusions so as to manipulate the object without significantly occluding it from view of the camera.

\subsection{Real World Benchmarking}
We follow the bin packing evaluation protocol used in ~\cite{lum2024dextrahgpixelstoactiondexterousarmhand, singh2025dextrahrgbvisuomotorpoliciesgrasp}. This benchmarking protocol involves placing 30 objects of varying size, shape, and weight on the table and having the policy grasp it. The goal of this is to assess the continuous performance of this task. We leverage the same state-machine as prior work. In this, the distilled policy also has an extra MLP head that predicts the position of the object. During rollout, when the object prediction head predicts the object to not be sufficiently lifted in the air, the state machine executes the policy's actions onto the real robot in order to grasp the object into the air. Once the predicted object position is sufficiently high, the state-machine transitions to a fixed motion to deposit the object in a bin before resetting the pose of the arm and running the policy again. The main metric we track is the success rate, which measures what percentage of objects were successfully grasped and deposited into the bin.

The results of this experiment are shown in Table~\ref{tab:model_sr}. The models trained from depth teachers outperformed all other models using state-based teachers, demonstrating that existing policies are often held back by the observability gap between the student and teachers. This shows end-to-end RL as a promising step towards improving robot policies. Furthermore, our policy with disaggregated simulation performed the best, which shows that increasing the batch size helps these vision policies. This is likely because tiled rendering environments are incredibly memory intensive which results in us saturating GPU memory before saturating PPO with a very large batch size.

\begin{table}[h]
  \centering
  \caption{Success rate by model.}
  \label{tab:model_sr}
  \begin{tabular*}{\columnwidth}{@{\extracolsep{\fill}}lc@{}}
    \toprule
    \textbf{Model} & \textbf{Success Rate $\uparrow$} \\
    \midrule
    DextrAH-G (state teacher) & $87\%$ \\[2pt]
    DextrAH-RGB (state teacher) & $77\%$ \\[2pt]
    Ours (depth teacher) & $87\%$ \\[2pt]
    Ours (depth teacher, disagg) & $\boldsymbol{93\%}$ \\
    \bottomrule
  \end{tabular*}
  \footnotetext[1]{SR = success rate.}
\end{table}

\section{Conclusion}
In this work, we demonstrate a method for end-to-end reinforcement learning with depth for dexterous grasping. Contrary to the standard paradigm of training state-based policies with RL and distilling them into vision-based ones, we are able to train depth policies with RL and distill them into RGB policies. We show that the lack of an information gap when distilling these depth policies results in better performance both in simulation and real-world experiment. We also present a method for efficiently training end-to-end RL policies with a low number of GPUs.

\bibliographystyle{IEEEtran}
\bibliography{references}

\end{document}